\documentclass[conference]{IEEEtran}
\usepackage{times}
\usepackage{url}

\usepackage[numbers]{natbib}
\usepackage{multicol}
\usepackage[bookmarks=true]{hyperref}
\usepackage{makecell}
\usepackage{graphicx}

\begin{document}

\title{Foresee: Attentive Future Projections of Chaotic Road Environments with Online Training}

\author{Anil Sharma and Prabhat Kumar
\\ 
Indraprastha Institute of Information Technology, Delhi \\
%
\{anils, prabhat17036\}@iiitd.ac.in}



%

\maketitle

\begin{abstract}
In this paper, we train a recurrent neural network to learn dynamics of a chaotic road environment and to project the future of the environment on an image. Future projection can be used to anticipate an unseen environment for example, in autonomous driving.  Road environment is highly dynamic and complex due to the interaction among traffic participants such as vehicles and pedestrians. 
Even in this complex environment, a human driver is efficacious to safely drive on chaotic roads irrespective of the number of traffic participants. The proliferation of deep learning research has shown the efficacy of neural networks in learning this human behavior. In the same direction, we investigate recurrent neural networks to understand the chaotic road environment which is shared by pedestrians, vehicles (cars, trucks, bicycles etc.), and sometimes animals as well. We propose \emph{Foresee}, a unidirectional gated recurrent units (GRUs) network with attention to project future of the environment in the form of images. 
We have collected several videos on Delhi roads consisting of various traffic participants, background and infrastructure differences (like 3D pedestrian crossing) at various times on various days. We train \emph{Foresee} in an unsupervised way and we use online training to project frames up to $0.5$ seconds in advance. We show that our proposed model performs better than state of the art methods (prednet~\cite{PredNet}, Enc. Dec. LSTM~\cite{srivastava}) and finally, we show that our trained model generalizes to a public dataset for future projections. 
\end{abstract}

\IEEEpeerreviewmaketitle

\section{Introduction}
\label{sec:intro}

\begin{figure}
\begin{center}
\includegraphics[width=3.3in]{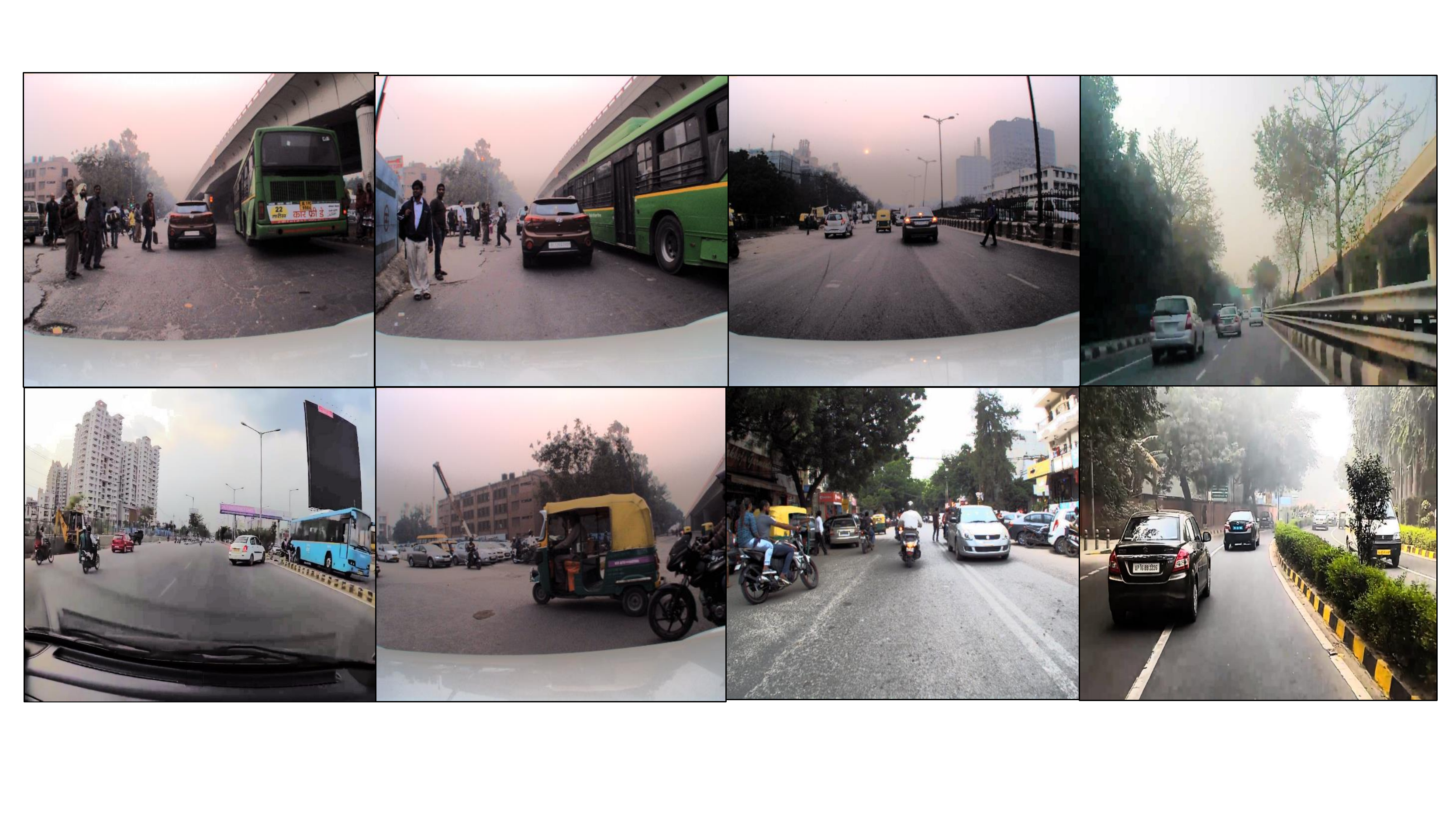}
\end{center}
\caption{Example scenarios showing chaotic movement of different kind of traffic participants on a typical road environment. Few images in first row shows that the pedestrians are randomly crossing vehicles and hinder vehicle movement even on the main road. Second row shows that the vehicles typically do not follow rules such as lane following and move in various directions making the anticipation task more difficult. In many images, lanes are also not visible.}
\label{fig:road_collage}
\end{figure}

Environment anticipation is an important task for situation awareness and decision making. There is recent progress in anticipation of road environments~\cite{ashesh,envanti} for safe driving and behavioral cloning~\cite{behclone} where an agent tries to clone behavior of a human driver. However, anticipation becomes difficult in real world because it is uncertain and dynamic~\cite{envprob}. Consider, for example, the road environment. The road environment is highly dynamic and stochastic due to the presence of a diverse set of human drivers and pedestrians, few examples are shown in figure~\ref{fig:road_collage}. The figure shows that the road space has chaotic movement of pedestrians and vehicles. We define chaotic environment as that environment where the traffic participants follow no rule and move randomly as shown in the figure. The same case is seen on road in developing countries like India. In such environments, the road space is shared by pedestrians, vehicles (cars, trucks, buses, motor-bikes etc.), and sometimes animals as well. Even when the environment is complex, its behavior can be modeled~\cite{ashesh}. Modeling such an environment requires detection, tracking and, understanding of the dynamics of the traffic participants. Given that they are also interacting with each other (for example, the lane change of one car on road affect the motion of other cars as well), the modelling is not trivial. Anticipating behavior of the environment is essential in various applications such as autonomous driving~\cite{shashuaAgent}, driving assistance~\cite{DPDM}, multi-target tracking~\cite{Reid}, autonomous landing on a moving target~\cite{landing}, etc. However, on the other hand, irrespective of the environment, humans are very good at anticipating such an environment. For example, they drive very successfully by anticipating maneuvers even in a very crowded chaotic shared space such as markets, street roads, highways etc. We explore ways to achieve that anticipation power in machines using neural networks by exploiting the predictive power of recurrent neural network to capture this human behavior. In this paper, we propose a deep learning architecture to generate future projections in terms of the camera frames few frames in advance. 
The future projections will help any robot/learning agent in situation awareness for decision making and planning in an unseen environment. 

In this work, we propose~\emph{Foresee}, a deep learning architecture for future projections of the chaotic road environment directly from the raw camera images. The network is composed of two layers of GRUs (Gated Recurrent Units~\cite{gru}) to encode the dynamics of the environment into a small representation in the hidden layers. Next, we reconstruct the future projections from the encoded representations using a fully connected layer. We train the network in an unsupervised way to achieve the desired performance. We formulate the above problem as a sequence generation task, where a sequence is the collection of images that are contiguous in time. We are interested in predicting the future from the past few sequence of frames. 
When we use \emph{Foresee} with online training, we are able to project the future up to $0.5$ seconds in advance.

Our specific research contributions are as follows:

\begin{enumerate}
    \item  We propose a deep learning architecture, \emph{Foresee}, using Gated Recurrent Units (GRUs) and attention for future projections. We show that the proposed architecture performs better than the current state of the art. We evaluate our proposal on a vast set of images collected in chaotic road environments of Indian roads. We investigate various design choices by  analyzing various hyper-parameters in \emph{Foresee}.
    \item We have collected a very large real road environment data using a monocular camera and dashcam videos from YouTube. In total, we have $101$ videos. The videos capture interaction of many traffic participants on various kinds of roads. For example, during urban driving, highway driving, merging at intersections, market, streets, etc. 
    
    \item We investigate and compare performance of two future projection architectures with our method on a very large dataset. 
    \item We explore online training on \emph{Foresee} to make projections. We observed that online training improves the performance and helps \emph{Foresee} to project future up to $0.5$ seconds in advance. Finally, we will show that \emph{Foresee} trained on our dataset generalizes to a public dataset. We will also show that the projected images can be used for steering angle estimation for behavioral cloning~\cite{behclone} in an autonomous driving simulator.
    
\end{enumerate}

The subsequent sections are structured as follows. Section~\ref{sec:related} describe state of the art for future predictions. In section~\ref{sec:methodology}, we provide details of the proposed system. Section~\ref{sec:experiments} describe the collected dataset, experimental setup and evaluation pipeline. In section~\ref{sec:results}, we demonstrate the future prediction results on the dataset collected on chaotic road environment. Section~\ref{sec:disc} has discussion and future work of our paper and section~\ref{sec:concl} concludes the paper.

\section{Related Works}
\label{sec:related}
In this section, we describe the state of the art approaches for environment anticipation. 

We are not first to look in this direction and related works have also explored future predictions from various viewpoints. One common approach is Bayesian filtering to predict next state as in Kalman filter~\cite{Kalman}. We include works that use neural networks as a learning architecture can directly anticipate environment from data without any explicit need of modeling.  Works like~\cite{yilmaz,fan2010human,deepTracking} have looked at Bayesian filtering for state prediction using neural networks. However, this was looked into separate blocks of object detection, tracking and prediction. Handcrafted features are used in such approaches which is an extra overhead. Authors in~\citet{YOLO,ROLO} have used CNN and LSTMs to find the future trajectory of an object using current camera location. They have first predicted the target location and then tracked it using LSTMs. The above approaches are supervised and requires a labeled dataset to predict target locations. The supervised learning for chaotic environments is very difficult as the environment is shared by different kinds of participants and a proper label for object locations is difficult to get. The labeling of images is a very costly task. 

Hence, an end-to-end learning approach is desired even for such a complex problem. One would also like an unsupervised learning task for a learning problem. The neural networks can capture the representative features for object patterns and motion dynamics. These can also capture the interactions among the traffic participants. For example, ~\cite{SocialLSTM} has modeled the interaction among pedestrians using a LSTM~\cite{lstm} network. In this, authors have looked upon the task of interactions among pedestrians as a social force model~\citet{social}. They have proposed a deep neural network model by using a separate LSTM model for each object and then the interactions are captured using a pooling layer among neighbors. Authors in ~\citet{deepTracking} have looked one step ahead for object tracking in partially observable environment. Their approach to bayesian filtering is end-to-end trainable and is unsupervised to predict the fully observable state. However, they test their approach on a simulation environment where dynamics are pre-defined and hence do not capture the chaotic environment. 

The papers~\citet{srivastava,PredNet} are very similar to our work. They show that their framework can predict future frame in advance using unsupervised learning. Authors in~\citet{srivastava} have used multilayer LSTM networks for future predictions and modeled it as a sequence prediction task. 
Whereas in~\citet{PredNet}, authors have proposed a video sequence prediction architecture using LSTM and predictive error coding. They evaluated their proposal on simple environment like fewer vehicles and only vehicles in the scene. We compare the performance of these methods with our proposed network on the dataset that we have collected for chaotic road environments. We observe that approach in~\cite{srivastava} fails beyond next frame and the approach in~\citet{PredNet} fails to encode the environment dynamics properly. It produces a high quality image which is more or less similar to previous frame. We compare and show that \emph{Foresee} performs better than above two for chaotic environments. The common problem with deep learning architectures is that they fail to generalize on different datasets but we will show that \emph{Foresee} trained on our dataset generalizes to kitti dataset for future projections.

\section{Proposed Methodology}
\label{sec:methodology}
In this section, we will explain \emph{Foresee} in detail that helps to predict the future projections of the road environment.

\begin{figure}[t]
\begin{center}
\includegraphics[width=3in]{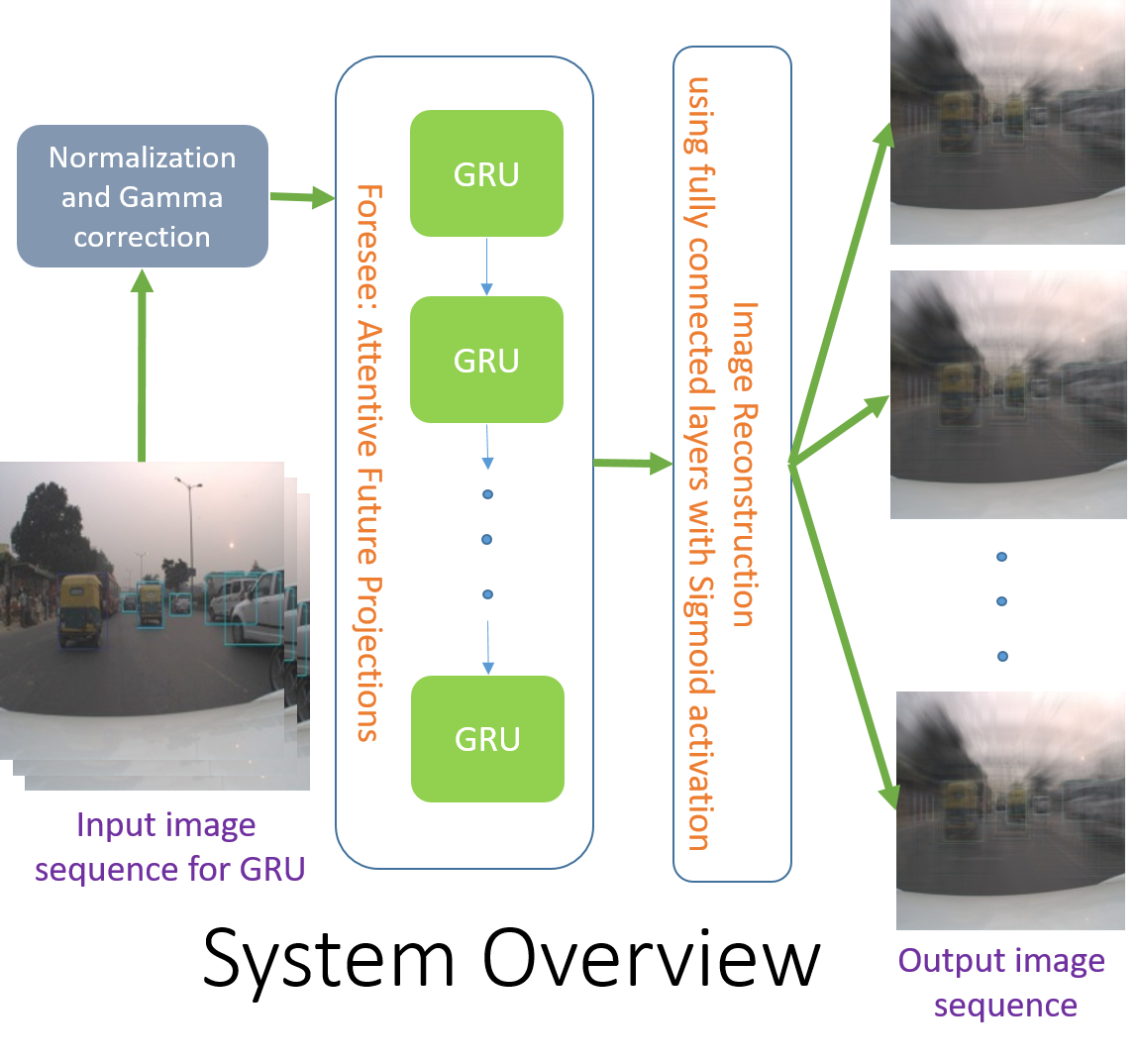}
\end{center}
\caption{System architecture for future projections.}
\label{fig:arch_new}
\end{figure}

\subsection{System Overview:}
In this subsection, we describe the system architecture diagrammed in figure~\ref{fig:arch_new}. 
The future projection is carried out in various steps. The input image is first normalized in range between $0$ and $1$ and then gamma correction is applied to enhance illumination. The normalized corrected image is then re-sized to shape $32*32*3$. An image sequence is then created by concatenating the last $10$ frames. The input image sequence is the sequence of images starting from current frame to $9$ frames in past and the output image sequence is the sequence of images in the future. The prepared image sequence is then passed to the recurrent network to encode the temporal sequence for future projections. The output sequence is reconstructed from the encoded representations using a fully connected layer with hidden units equal to the number of pixels in the output image ($32*32*3$). In the next subsection, we will explain the recurrent network named \emph{Foresee}, which is composed of stacked GRU cells which are used recursively to predict future projections. 

\subsection{Foresee: Recursive Future projections using GRU Network and Attention}

\begin{figure*}[h]
\begin{center}
\includegraphics[width=5.5in]{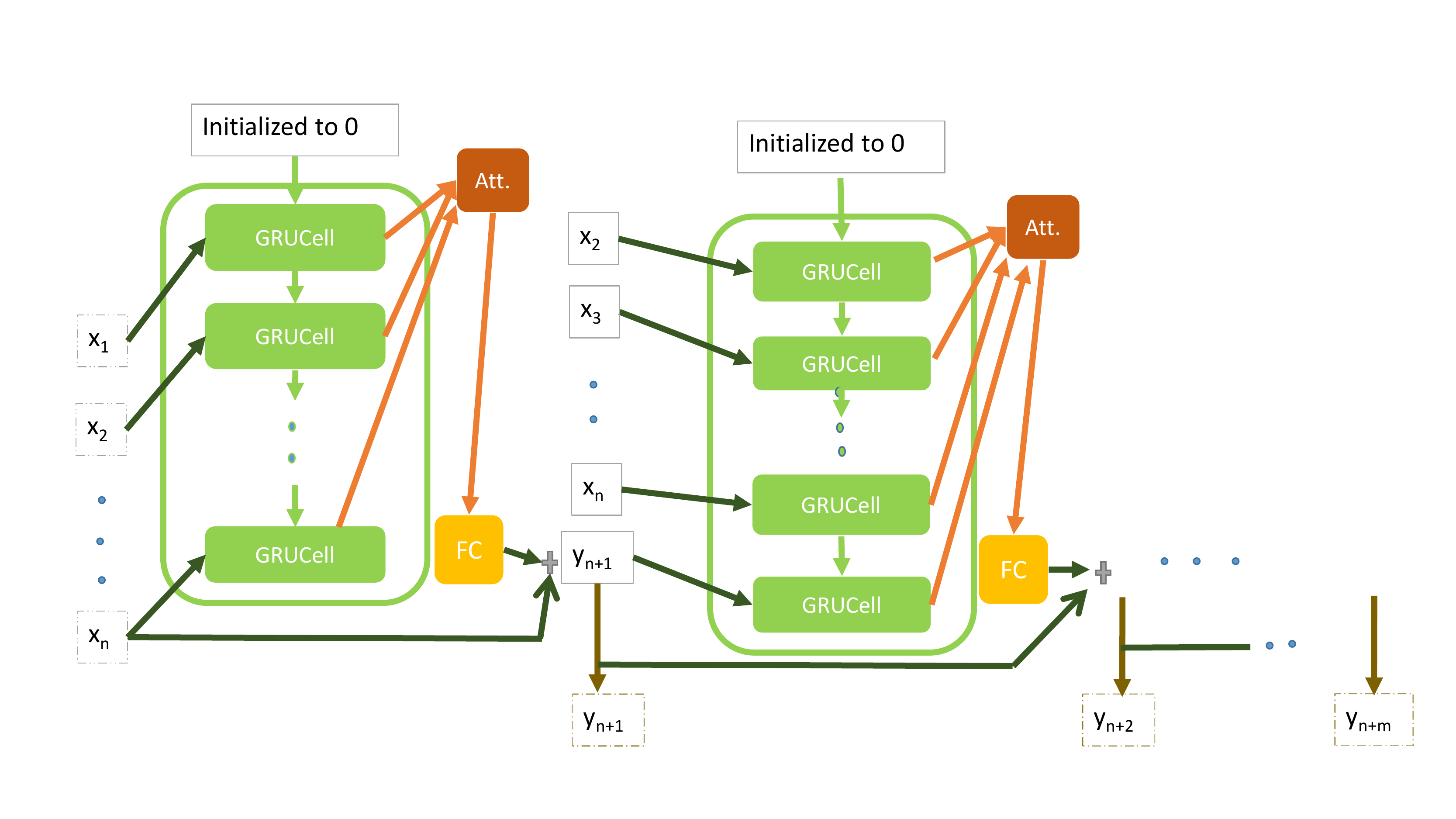}
\end{center}
\caption{Foresee model: attention is applied on the GRUCells. Single block is foresee and it is applied recursively to predict a longer output sequence.}
\label{fig:GRUAtt}
\end{figure*}

To predict the future projections, a stack of GRU cells~\cite{gru} is employed along with attention~\cite{colin,Bahdanau}. In the subsequent text, we will explain GRU and attention method. 
The Foresee network consists of GRUCells which has a hidden state corresponding to each time step. Refer to figure~\ref{fig:GRUAtt}, one block of the figure is \emph{Foresee}. The figure diagrams the recursive use of \emph{Foresee} for future projections few frames in advance. 

A Gated Recurrent Unit (GRU) cell is the slight variation of the Long Short Term Memory (LSTM) Cell where the forget and input gate of the LSTM is combined into a single update gate. It also combines the cell state and the hidden state into the hidden state itself. For more details on the LSTM, the readers can refer to online tutorial~\footnote{http://colah.github.io/posts/2015-08-Understanding-LSTMs/}. The input $x_t$ at time $t$ is fed into the network and the necessary information to encode the temporal sequence till time $t$ is stored in the hidden state of the GRU cell. For each frame in the input sequence, each layer computes following functions:

\begin{equation}
r_t = \sigma(W_{ir}*x_t + b_{ir} + W_{hr}*h_{t-1} + b_{hr})    
\end{equation}
\begin{equation}
z_t = \sigma(W_{iz}*x_t + b_{iz} + W_{hz}*h_{t-1} + b_{hz})    
\end{equation}

\begin{equation}
n_t = Tanh(W_{in}*x_t + b_{in} + r_t*(W_{hn}*h_{t-1}+b_{hn}) + b_{hn})    
\end{equation}
\begin{equation}
\label{eq:gruHidden}
h_t = (1-z_t)*n_t + z_t*h_{t-1}
\end{equation}

where $h_t$ is the hidden state at time $t$, $x_t$ is the output of previous layer at time $t$ or the input at time $t$ for the first layer. $r_t$ is the reset gate, $n_t$ is the new gate and $z_t$ is the update gate. The $W_{i}$'s are the weight parameters for the $i^{th}$ gate. 

\begin{figure}[h]
\begin{center}
\includegraphics[width=2.8in]{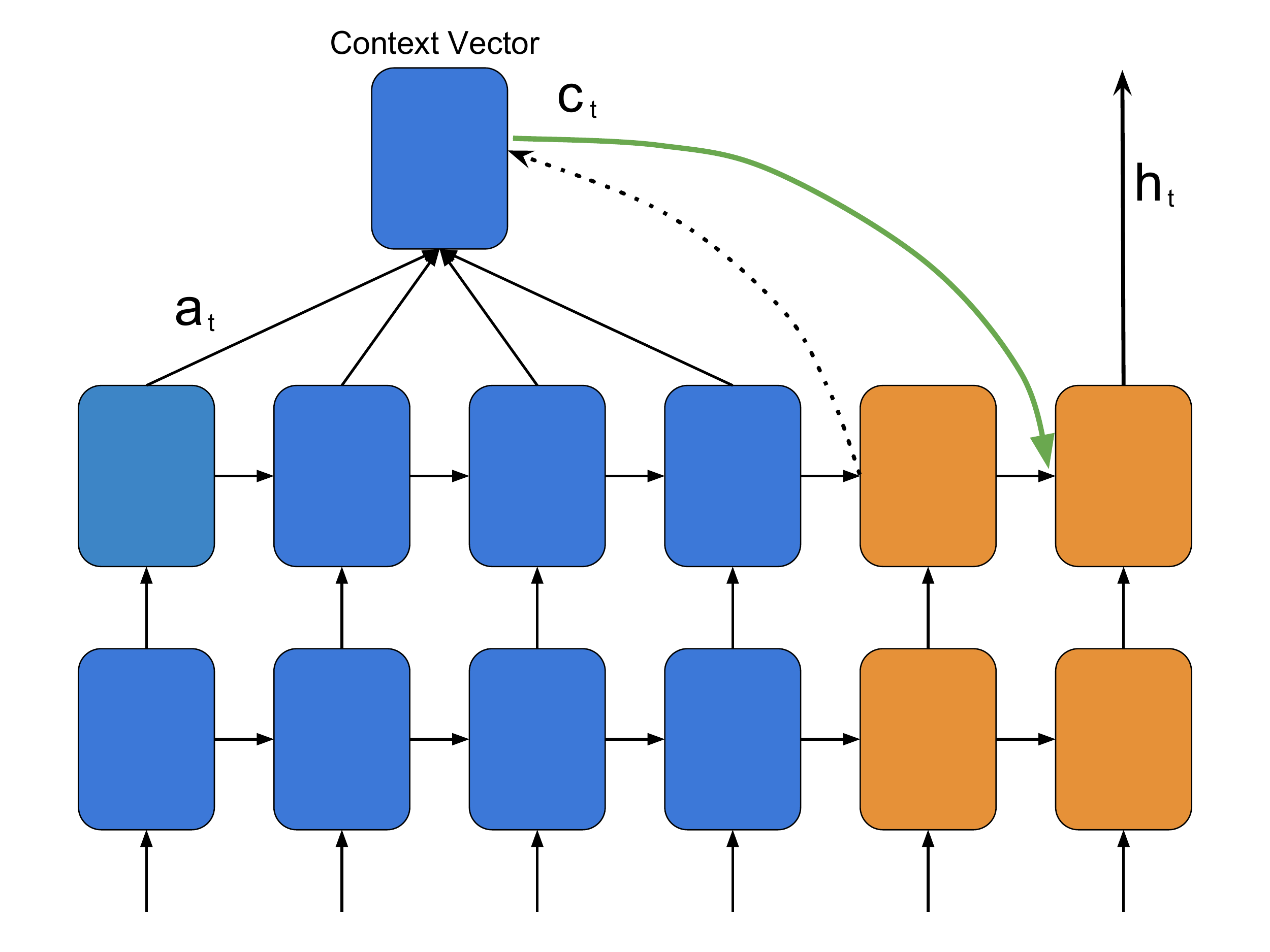}
\end{center}
\caption{Attention mechanism to improve the quality at a larger time step. In conventional GRUs, the quality degrades for longer sequences.}
\label{fig:attention}
\end{figure}

In GRU networks, the reconstruction quality degrades with the longer sequence as it cannot stuff all the information into its hidden layer (see ~\cite{Bahdanau}). To resolve this problem, attention methods were employed, for example, in text generation~\cite{Bahdanau}, this was employed to improve long term dependency. We observed that using attention, the output is not only the function of previous time hidden-state and current input but it is now a function which computes the weighted sum of all input encodings. Since road environment is not markov i.e., it does not depend only on previous frame, the attention method helps the network to attend to past frames as compared to only the previous frame.  Figure~\ref{fig:attention}
shows a graphical representation of the attention (weighted sum using a context vector). The attention mechanism takes outputs from all previous time steps and makes a context vector which is a weighted sum of the representations at previous time steps. The context vector is then multiplied with the new hidden state and then the output is reconstructed. The attention layer performs following operations on the hidden state of the GRU network(equation~\ref{eq:gruHidden}).

\begin{equation}
e_{ij} = Tanh(Mul(O_t,W) + b)
\end{equation}
\begin{equation}
a_i = exp(e_{ij})
\end{equation}
\begin{equation}
C_t = Softmax(a_i)
\end{equation}
\begin{equation}
O_{t}^{wt} = Mul(O_t,C_t)
\end{equation}

where $O_t$ is the output sequence (outputs for all timestamps) of the GRU network, $C_t$ is the attention context vector, $Mul$ is the matrix multiplication function, $O_{t}^{wt}$ is the weighted output at time $t$. The training loss is mean squared error between target frame $T_t$ and projected frame $O_{t}^{wt}$. The training loss is mentioned in equation~\ref{eq:loss}.

\begin{equation}
\label{eq:loss}
L_{train} = \frac{1}{N} \sum_{i=0}^{N} |T_t(i)-O_{t}^{wt}(i)|^2
\end{equation}

The above procedure is applied to project the next frame from the input sequence (please note that we are using input sequence of length $10$). We investigate the different ways of attention and other hyper-parameters in section~\ref{sec:understandForesee}. To generate projections of many frames \emph{Foresee} is used recursively. The recursive \emph{Foresee} is detailed in figure~\ref{fig:GRUAtt}.

\section{Experimental Setup}
\label{sec:experiments}
In this section, we demonstrate the dataset, experimental setup and the performance metric used for the evaluation of \emph{Foresee} and its comparison with state of the art methods.

\emph{Dataset}:  To capture the chaotic environment, we collected data on Delhi roads where the road space is shared by pedestrians, vehicles (cars, buses, auto-rickshaws, bicycles etc), and sometime animals as well. We have collected data for urban traffic and highway traffic on Delhi roads under both chaotic and ordered traffic situations. We have collected real road environment images using
a car-mounted Point-Grey monocular camera on various times on various days on Delhi roads. The FPS was varying but the videos
are then standardized to $10$ FPS. In addition to the data collection on Delhi roads, we have made use of dash-cam videos available on YouTube. We selected videos of the chaotic situations (defined in introduction) only and were
from various states in India. In total, we have $101$ videos. These videos are then randomly splitted into three sets for training, validation and testing. The training set is used to train the model, validation set is used to decide whether the network is over-fitting and testing set is used to evaluate the trained model and for performance comparison with other methods. All the sets are sufficiently large and cater to various environmental settings. The dataset description is given in table~\ref{tab:datadescr}. Just to say, the dataset was collected on the wild and contain environments like market, heavy traffic, mild traffic, mixed objects (pedestrians, vehicles, animals etc.), different backgrounds (building, trees etc.) and different infrastructural variations (3D pedestrian crossing etc.). The intuition behind collection of such a dataset is to develop a deep learning architecture which can be generalized to various real road environments. The data frames were normalized between $0$ and $1$ using opencv and re-sized to $32*32*3$. We choose a smaller image size because with smaller image size the network is less complex and also deep learning is able to make sense of the objects and environment on smaller images, for example, cifar dataset~\cite{cifar} has $32*32*3$ size images and is widely used for classification. 

\begin{table}[t]
\caption{Dataset description}
\label{tab:datadescr}
\begin{center}
\begin{tabular}{lll}
\multicolumn{1}{c}{\bf Set}  &\multicolumn{1}{c}{\bf Number of videos}
&\multicolumn{1}{c}{\bf Number of images}
\\ \hline \\
Training &55 &82,265 \\
Validation &22 &4,314 \\
Testing &24 &14,500 \\
Total &101 &101,079 \\
\hline
\end{tabular}
\end{center}
\end{table}

\emph{Evaluation Metric}:To quantitatively assess the projected image quality, we have used mean square error (MSE) and structural similarity index measure (SSIM)~\cite{ssim} as the evaluation metric. MSE at time $t$ is the mean square error between target image at time $t$ and projected image at time $t$. SSIM assess the image quality based on the structural degradation and compares an image with a reference image.

\emph{Experiments}: We designed following experiments for better understanding and evaluation of the proposed architectures for future projections: 


\begin{enumerate}
    \item Qualitative understanding of \emph{Foresee} and what representations it is able to learn
    
    \item Quantitative understanding of  \emph{Foresee} and its hyper-parameters
    
    \item Performance comparison with state of the art methods for future projections. We compare with encoder-decoder LSTM method~\cite{srivastava} and prednet~\cite{PredNet}.
    
    \item \emph{Foresee} with online training to see the benefits online training can provide for the future projections
    
    \item Quantitative evaluation of \emph{Foresee} on Kitti dataset~\cite{kitti} to check the generalization power of \emph{Foresee}.
    
    \item Steering estimation on an autonomous driving simulator to check how well the projection help for behavioral cloning. 
    

\end{enumerate}

\emph{Implementation Details:} Foresee is implemented in pytorch with gradient computation in adagrad. The whole computation is done on a Tesla K20m GPU.

\section{Results}
\label{sec:results}
In this section, we will show the efficacy of our proposed framework using the experiments mentioned in the previous section. 



\subsection{Qualitative understanding of \emph{Foresee} and what representations it is able to learn}

\begin{figure}
\begin{center}
\includegraphics[width=2.5in]{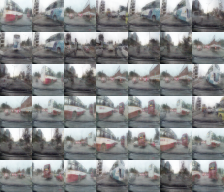}
\end{center}
\caption{Image showing next frame projection using \emph{Foresee}. All images are generated using \emph{Foresee}. Each image is $32*32*3$. } 
\label{fig:outseq_foresee}
\end{figure}

\begin{figure*}
\begin{center}
\includegraphics[width=5.5in]{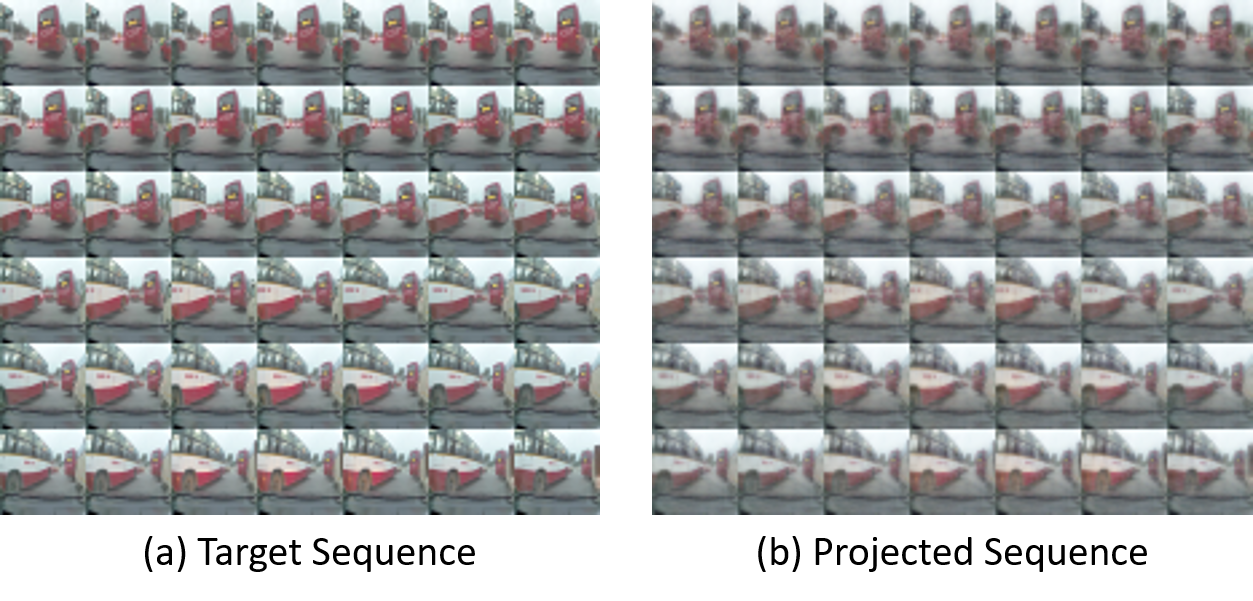}
\end{center}
\caption{Image showing (a) target sequence of images from the test set, (b) corresponding next frame projections generated using \emph{Foresee}. } 
\label{fig:proj_seq}
\end{figure*}

First of all, 
we investigated various sequence prediction architectures as explained in Andreas Karpathys blog \footnote{http://karpathy.github.io/2015/05/21/rnn-effectiveness/}. We found many-to-many sequence prediction networks to be more effective for encoding the environment representations for future projections (see section~\ref{sec:understandForesee}). Method proposed in~\cite{srivastava} is also a many-to-many sequence prediction network using LSTMs (Long Short Term Memory~\cite{lstm}). 
In the many-to-many architectures, we first employed the approach proposed by Srivastava et. al.~\cite{srivastava} and identified that it is not able to persist the sequence representation even for few frames because the road environment is continuously changing and the next frame does not depend completely on the current frame, we need a method to attend previous frames as well. For this, we investigated various approaches such as reconstruction mechanisms using deconvolution layers and fully connected layers, feature representation using RESNET~\cite{resnet} and attention~\cite{colin}. We observed that attention is performing the best for the chaotic environments. In the same architecture, GRUs are performing better than LSTMs. For brevity, we show the future projections only for GRU with attention (which is \emph{Foresee}) in figure~\ref{fig:outseq_foresee}. For choosing hyper-parameters of the network, we did an exhaustive search over multiple hidden state sizes and input sequence lengths, etc (refer section~\ref{sec:understandForesee}). In the final network, the hidden state size is $512$ and the input sequence length is $10$ frames ($1$ second). Our model has $2$ layers of GRU cells. The learning rate and the weight optimization algorithm impacted the performance a lot. We observed highest performance when using the Adam algorithm~\cite{adam} for weight optimization. 
We try \emph{Foresee+online} in which we first apply online training to the input sequence and then decode the future projections. Out of all tested models, we show results of \emph{Foresee} on our test set in figure~\ref{fig:outseq_foresee} and \emph{Foresee+online} in figure~\ref{fig:seq_foreseeonline}. Figure~\ref{fig:proj_seq} show projected image sequence of a video from test set along with the target image sequence. 

\begin{table}[h]
\caption{Training on different many-to-many sequence prediction architectures. MM-2 is the many-to-many sequence when input and output sequences are synced. MM-1 is the architecture when output length is larger than input length. These architecture are taken from~\cite{andreas}. }
\label{tab:seqpred}
\begin{center}
\begin{tabular}{lll}
\hline 
\multicolumn{1}{c}{\bf Approaches} 
&\multicolumn{1}{c}{\bf \makecell{MSE(train)\\ $10^{-4}$}} 
&\multicolumn{1}{c}{\bf \makecell{MSE(val)\\ $10^{-4}$}} 
\\ \hline \\
MM-1 (output)  &$0.52$ & $2.7$ \\
MM-2 (output)  &$0.40$ & $0.85$ \\
\hline
\end{tabular}
\end{center}
\end{table}


\begin{table*}[h]
\caption{Performance with various hyper-parameters when attention is applied at the hidden layer (AttnHidden) and the reconstructed output (AttnOutput). MSE(Last) show the MSE when the attention is applied only at the last time step during the training procedure and MSE(All) signifies the MSE value when the attention is applied at all steps of encoding and decoding.}
\label{tab:paramAttnAll}
\begin{center}
\begin{tabular}{lllll}
\hline
\multicolumn{1}{c}{\bf \makecell{hyper-param\\(Input,hidden) } } 
&\multicolumn{2}{c}{\bf \makecell{AttnHidden} } 
&\multicolumn{2}{c}{\bf \makecell{AttnOutput} } \\
\hline

&\multicolumn{1}{c}{\bf \makecell{MSE (Last) \\ $10^{-4}$} } 
&\multicolumn{1}{c}{\bf \makecell{MSE (All) \\ $10^{-4}$}} 
&\multicolumn{1}{c}{\bf \makecell{MSE (Last) \\ $10^{-4}$} } 
&\multicolumn{1}{c}{\bf \makecell{MSE (All) \\ $10^{-4}$}} 
\\ \hline 
10, 512  &$3.2$ & $2.9$ &$0.67$ & {\bf 0.00189} \\
20, 512  &$2.9$& $3.5$ &$2.7$& $2.2$ \\
10, 1024 &$4.5$& $7$  &$11$& $1.1$ \\
20, 1024 &$13$& $2.4$  &$2.1$& $0.85$ \\
\hline
\end{tabular}
\end{center}
\end{table*}

\subsection{Quantitative understanding of  \emph{Foresee} and its hyper-parameters}
\label{sec:understandForesee}

\begin{figure}
\begin{center}
\includegraphics[width=2.2in]{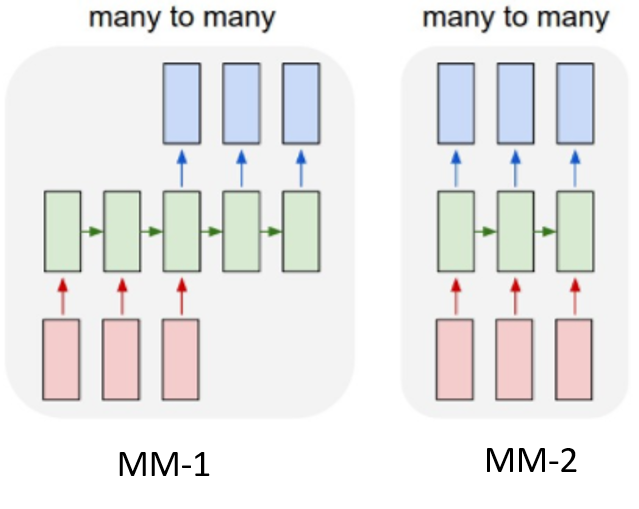}
\end{center}
\caption{Many-to-many sequence generation in recurrent networks. }

\label{fig:andreas_mm}
\end{figure}

\emph{Foresee} has various hyper-parameters that require tuning to achieve better results. The investigated parameters are input sequence length, hidden size of GRU cell, attention mechanism and training procedure. Table~\ref{tab:paramAttnAll} shows the average mean square error on the validation set using different combinations of hyper-parameters when the attention is used at the topmost hidden layer and at the reconstructed output. Analysis is following: 

\begin{enumerate}
\item Input sequence length (named Input in table~\ref{tab:paramAttnAll}): We expect \emph{Foresee} to project future up to $0.5$ seconds in advance. For this we explored input length of $1,2$ seconds. We didn't go beyond $2$ seconds because it is clear from the table that the model started over-fitting. This is expected as the road environment is dynamic.  
\item Hidden Size of GRU cell (hidden): We investigated various hidden sizes. For an input image of size $32*32*3$ ($3072$ pixels), we tried hidden size of $512$ and $1024$. Refer to table~\ref{tab:paramAttnAll} which shows performance of different hidden sizes, we observe that network overfits for larger hidden sizes other than the last column. Last column shows that the attention is applied at reconstructed output and at all time steps. Hidden size of $512$ and input sequence length of $10$ performed the best.
\item Attention mechanism: Attention helps the network to attend to a specific part of input sequence. Attention can be applied at several locations in the network. For example, at the reconstructed output (named AttnOutput) or at the hidden state (named AttnHidden). It can also be chosen when to apply it, while encoding, decoding or at both. Table~\ref{tab:paramAttnAll} shows the average MSE when attention is applied at topmost hidden layer (first two columns) and when it is applied at the reconstructed output (last two columns). MSE(Last) signifies the MSE when attention is applied only for the last time step whereas MSE(All) signifies that attention is applied at all steps of the input sequence along with decoder. Intuitively one would expect hidden to perform better because if the hidden state is not good enough then output will surely be worse but here attention at output performs better. The initial reconstructions are better and attention helped to make use of it. Whereas when attention is used at hidden a slight error at hidden will also propagate while reconstruction.
\item Training procedure: For many-to-many sequence generation, we used encoder-decoder training procedure. The procedure are shown in figure~\ref{fig:andreas_mm}. 
We tried to backpropagate with error computation only for encoder (named MM-2), only for decoder (MM-1). 
Table~\ref{tab:seqpred} shows the two sequence prediction architectures when attention is applied at output. MM-1 (output sequence is longer than input during training) is over-fitting and MM-2 when input and output sequence length are in sync performs better.The two networks are shown in figure~\ref{fig:andreas_mm}. 

\end{enumerate}

\subsection{Performance comparison with state of the art methods}

In this section, we compare performance of various future projection architectures with \emph{Foresee} and the performance is quantified using MSE and SSIM. Table~\ref{tab:perfcomp} shows the  MSE and SSIM values of the three architectures, approaches of ~\cite{srivastava} and~\cite{PredNet} and our proposed approach without and with online training. The results were computed on our test set when the model is trained with our train set. 
Copy last frame is the trivial approach when previous frame is used as the projected frame, in this case MSE is very high and since the image does not have noise and share the background it has high SSIM.  

Prednet~\cite{PredNet}: We trained prednet on our training set. We stopped the training process after a day of training and when the validation loss reached approx. $6*10^{-3}$. To produce results of prednet, we have used the scripts provided. The output of prednet is $128*160*3$ image. On the original projections, the MSE of prednet was $8.01*10^{-2}$ and SSIM was $47.18$. To make the comparison fair we downsampled prednet output and then compute MSE and SSIM which are shown in table~\ref{tab:perfcomp}. Prednet makes use of the previous frame and then performs error correction. In highly dynamic environment the prednet returns only the previous frame and doesn't encode the motion dynamics into the representations and hence the output is very bad both visually and quantitatively. 


We have generated all possible sequence of length $10$ starting from first frame on the testing set. The results shown in the table are based on the exhaustive set of sequences.

Enc. Dec. LSTM~\cite{srivastava}: We trained their architecture on our training with $1024$ hidden size and input sequence length of $10$ frames (they reported same parameters in their paper on a $32*32$ image patch). As noted, a sequence prediction architecture without attention does not encode environment representations. 

\begin{table}[t]
\caption{Performance comparison on test data.}
\label{tab:perfcomp}
\begin{center}
\begin{tabular}{lll}
\hline
\multicolumn{1}{c}{\bf Approach} 
&\multicolumn{1}{c}{\bf MSE} &\multicolumn{1}{c}{\bf SSIM}
\\ \hline \\
Enc. Dec. LSTM  &$7.9*10^{-4}$ & $62.92$ \\
PredNet  &$5.8*10^{ -2}$ & $70.04$ \\
Copy last frame &$2.1*10^{-3}$ & $93.40$ \\
Foresee  &$1.08*10^{-5}$ & $86.40$ \\
Foresee with online  &$7.7*10^{-6}$ & $86.43$  \\
\hline
\end{tabular}
\end{center}
\end{table}


\subsection{\emph{Foresee} with online training}


At any point of time, the agent/vehicle has access to the previous frames that share the same background as the current frame or few frames in future and hence these can be used for further training. 
By this way, \emph{Foresee} will get to know about the background of the input images which it can adapt to and can project the better representations. Since averaging with previous frame improves results, we averaged the input at the GRUcell with the previous frame at every time steps instead of averaging on the output sequence. In brief, online training is the real-time training of the network on the previously seen frames. 

\begin{figure}
\begin{center}
\includegraphics[width=3in]{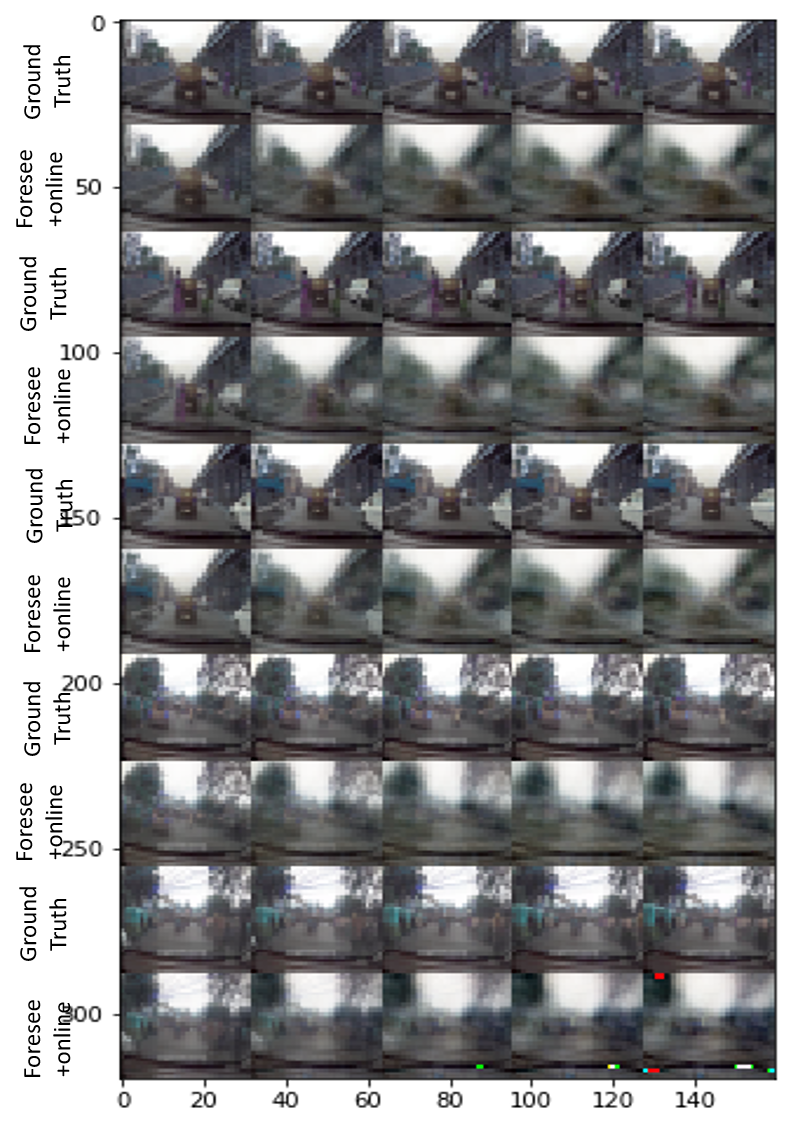}
\end{center}
\caption{Future projection for up to $5$ frames using \emph{Foresee+online}.}

\label{fig:seq_foreseeonline}
\end{figure}

For online training, we used previously available frames (up to $1$ second) to train the model again using Adam as the optimizer. Next, we projected $5$ frames ($0.5$ second) using the newly trained model. Previously trained network (on training set) is used from scratch for each video. Using online training, \emph{Foresee} is able to predict frame $0.5$ seconds in advance with an average MSE of $4.02*10^{-5}$ on our test set with SSIM on first frame of $86.43$. The table~\ref{tab:perfcomp} shows error values for the next frame projection only. 
Figure~\ref{fig:seq_foreseeonline} shows the projected output for $0.5$ seconds. The network is able to preserve the representation for longer than $1$ frame. 




\subsection{Quantitative evaluation of \emph{Foresee} on Kitti dataset}

\begin{figure}[h]
\begin{center}
\includegraphics[width=3in]{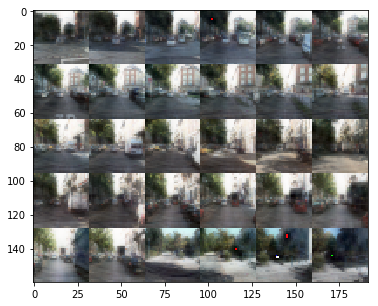}
\end{center}
\caption{Future projections on Kitti dataset when \emph{Foresee} is trained on our training dataset.}

\label{fig:kitti_online}
\end{figure}

In this section, we show the generalization power of our model on a public dataset. We used the previously trained model (on our training set) and test future projections for Kitti dataset~\cite{kitti} which is also used in prednet~\cite{PredNet}. With \emph{Foresee+online}, we achieve $4.7*10^{-4}$ MSE and $76.76$ SSIM. Figure~\ref{fig:kitti_online} shows the projected images using \emph{Foresee+online}. The figure clearly shows that \emph{Foresee+online} is able to learn representations on another data set and it generalizes to other environments. 

\subsection{Steering Estimation on the projected images:}

\begin{figure}
\begin{center}
\includegraphics[width=3.3in]{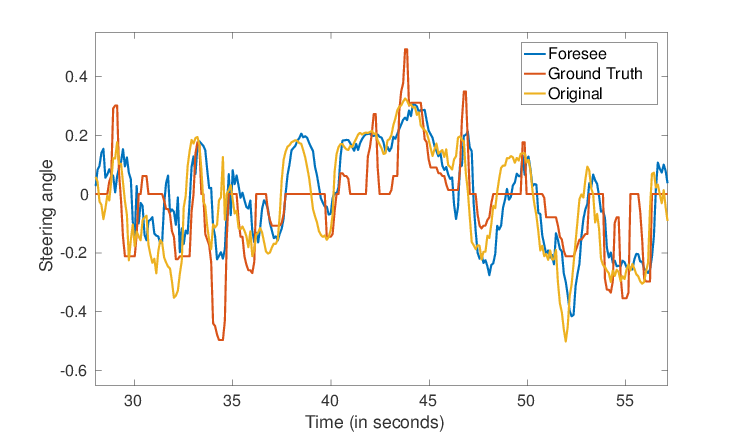}
\end{center}
\caption{Figure showing steering angle values estimated from the projected images (\emph{Foresee}), Ground truth angle and steering angle estimated from the original images (Original in figure legend). }

\label{fig:steering}
\end{figure}

We use projected image from \emph{Foresee} for behavioral cloning~\cite{steering, behclone}. In Behavioral cloning~\cite{steering, behclone}, we train a neural network to capture the behavior of a human driver for an autonomous vehicle. The trained model will try to estimate steering angle from images for lane keeping. Estimating steering angles from future images may help to avoid a near miss or improve drive. We use images from an autonomous vehicle simulator~\cite{behclonedata} for steering angle estimation. The available data was divided into $60\%$ training and $40\%$ testing after steering angle normalization. We trained a convolution neural network with MSE loss on simulator training set. We did not train \emph{Foresee} again on these images and generated future projections from the previously trained model (on our train set). The projected images are then used for steering angle estimation. The average MSE in the steering angle values estimated from \emph{Foresee} is $0.0164$ $degree^2$ and $0.0172$ $degree^2$ when estimated using original images. The estimated values from the network are shown in figure~\ref{fig:steering}. The figure shows estimation from \emph{Foresee} generated images, original images and ground truth values.


\section{Discussion}
\label{sec:disc}
Future projections are essential for behavioral cloning~\cite{behclone} in an unseen environment. We showed that GRU with attention is able to achieve better performance for future projections as compared to other available methods. The task of future projection has been looked in unsupervised way in the literature. However, no work has explored online training for such a task as the environment is repeating and online training can help in better projections. We showed a trivial online training approach which improves the error rate and helps in projection of a longer sequence. More online training methods can be explored for improvements in this direction. The future projections can be used other tasks such as tracking, cooperative perception and behavioral cloning. 

\section{Conclusion}
\label{sec:concl}
We proposed \emph{Foresee}, a deep learning architecture for future projections using Gated Recurrent Units and attention methods. We showed that attention when applied at all steps of the reconstructed output of the input sequence performs better. We collected a very large data set of chaotic road environments which capture different traffic participants and different infrastructural variations. 
We showed that the proposed architecture performs better than the state of the art methods for future projections. At end we showed that \emph{Foresee} with online training which is able to project future for up to $0.5$ seconds in advance and generalizes to a public dataset for road environments. The projected images were shown to be effective in steering estimation on simulated images from an autonomous vehicle simulator.

\section*{Acknowledgments}
The authors would like to thank team Swarath\footnote{\url{https://www.facebook.com/swarathatiiitd/}} at IIIT-Delhi for their help in data collection. The authors would also like to thank Dr. Arun Balaji Buduru, Assistant Professor at IIIT-Delhi, for his value able comments and discussions. 


\bibliographystyle{plainnat}
\bibliography{references}

\end{document}